\documentclass{article}

\usepackage[english]{babel}

\usepackage{times}
\usepackage{graphicx} 
\usepackage{subfigure} 

\usepackage{natbib}

\usepackage{algorithm}
\usepackage{algorithmic}

\usepackage{hyperref}

\usepackage[accepted]{icml2017}

\icmltitlerunning{Tensor-Train Recurrent Neural Networks for Video Classification}

\usepackage{subfigure}
\usepackage{widetext}

\usepackage{makeidx}
\usepackage[english]{babel}
\usepackage[utf8]{inputenc}
\usepackage{amsmath}
\usepackage{amsfonts}
\usepackage{varwidth}
\usepackage{multirow}
\usepackage[table]{xcolor}

\usepackage{graphicx}
\usepackage{caption}

\usepackage{natbib}

\usepackage{algorithm}
\usepackage{algorithmic}

\newcommand{\bs}[1]{{\boldsymbol{#1}}}

\begin{document} 

\twocolumn[
\icmltitle{Tensor-Train Recurrent Neural Networks for Video Classification}




\begin{icmlauthorlist}
\icmlauthor{Yinchong Yang}{lmu,siemens}
\icmlauthor{Denis Krompass}{siemens}
\icmlauthor{Volker Tresp}{lmu,siemens}
\end{icmlauthorlist}

\icmlaffiliation{lmu}{Ludwig Maximilian University of Munich, Germany}
\icmlaffiliation{siemens}{Siemens AG, Corporate Technology, Germany}

\icmlcorrespondingauthor{Yinchong Yang}{yinchong.yang@siemens.com}

\icmlkeywords{boring formatting information, machine learning, ICML}

\vskip 0.3in
]



\printAffiliationsAndNotice{}  

\begin{abstract}

The Recurrent Neural Networks and their variants have shown promising performances in sequence modeling tasks such as Natural Language Processing. 
These models, however, turn out to be impractical and difficult to train when exposed to very high-dimensional inputs due to the large input-to-hidden weight matrix. 
This may have prevented RNNs' large-scale application in tasks that involve very high input dimensions such as video modeling; current approaches reduce the input dimensions using various feature extractors. 
To address this challenge, we propose a new, more general and efficient approach by factorizing the input-to-hidden weight matrix using Tensor-Train decomposition which is trained simultaneously with the weights themselves. 
We test our model on classification tasks using multiple real-world video datasets and achieve competitive performances with state-of-the-art models, even though our model architecture is orders of magnitude less complex. 
We believe that the proposed approach provides a novel and fundamental building block for modeling high-dimensional sequential data with RNN architectures and opens up many possibilities to transfer the expressive and advanced architectures from other domains such as NLP to modeling high-dimensional sequential data. 
\end{abstract}

\section{Introduction}
Nowadays, the Recurrent Neural Network (RNN), especially its more advanced variants such as the LSTM and the GRU, belong to the most successful machine learning approaches when it comes to sequence modeling. Especially in Natural Language Processing (NLP), great improvements have been achieved by exploiting these Neural Network architectures. This success motivates efforts to also apply these RNNs to video data, since a video clip could be seen as a sequence of image frames. 
However, plain RNN models turn out to be impractical and difficult to train directly on video data due to the fact that each image frame typically forms a relatively high-dimensional input, which makes the weight matrix mapping from the input to the hidden layer in RNNs extremely large. For instance, in case of an RGB video clip with a frame size of say 160$\times$120$\times$3, the input vector for the RNN would already be $57,600$ at each time step. In this case, even a small hidden layer consisting of only 100 hidden nodes would lead to 5,760,000 free parameters, only considering the input-to-hidden mapping in the model. 

In order to circumvent this problem, state-of-the-art approaches often involve pre-processing each frame using Convolution Neural Networks (CNN), a Neural Network model proven to be most successful in image modeling. The CNNs do not only reduce the input dimension, but can also generate more compact and informative representations that serve as input to the RNN.  
Intuitive and tempting as it is, training such a model from scratch in an end-to-end fashion turns out to be impractical for large video datasets. Thus, many current works following this concept focus on the CNN part and reduce the size of RNN in term of sequence length \cite{donahue2015long, sr2015uns}, while other works exploit pre-trained deep CNNs as pre-processor to generate static features as input to RNNs \cite{yue2015beyond, donahue2015long, sharma2015action}. The former approach neglects the capability of RNNs to handle sequences of variable lengths and therefore does not scale to larger, more realistic video data. The second approach might suffer from suboptimal weight parameters by not being trained end-to-end \cite{fernando2016learning}. Furthermore, since these CNNs are pre-trained on existing image datasets, it remains unclear how well the CNNs can generalize to video frames that could be of totally different nature from the image training sets. 

Alternative approaches were earlier applied to generate image representations using dimension reductions such as PCA \cite{zhang1997face, kambhatla1997dimension, ye2004gpca} and Random Projection \cite{bingham2001random}. Classifiers were built on such features to perform object and face recognition tasks. These models, however, are often restricted to be linear and cannot be trained jointly with the classifier. 

In this work, we pursue a new direction where the RNN is exposed to the raw pixels on each frame without any CNN being involved. 
At each time step, the RNN first maps the large pixel input to a latent vector in a typically much lower dimensional space. Recurrently, each latent vector is then enriched by its predecessor at the last time step with a hidden-to-hidden mapping. In this way, the RNN is expected to capture the inter-frame transition patterns to extract the representation for the entire sequence of frames, analogous to RNNs generating a sentence representation based on word embeddings in NLP  \cite{sutskever2014sequence}. In comparison with other mapping techniques, a direct input-to-hidden mapping in an RNN has several advantages. First it is much simpler to train than deep CNNs in an end-to-end fashion. Secondly it is exposed to the complete pixel input without the linear limitation as PCA and Random Projection. Thirdly and most importantly, since the input-to-hidden and hidden-to-hidden mappings are trained jointly, the RNN is expected to capture the correlation between spatial and temporal patterns.

To address the issue of having too large of a weight matrix for the input-to-hidden mapping in RNN models, we propose to factorize the matrix with the Tensor-Train decomposition \cite{oseledets2011tensor}. In \cite{novikov2015tensorizing} the Tensor-Train has been applied to factorize a fully-connected feed-forward layer that can consume image pixels as well as latent features. 
We conducted experiments on three large-scale video datasets that are popular benchmarks in the community, and give empirical proof that the proposed approach makes very simple RNN architectures competitive with the state-of-the-art models, even though they are of several orders of magnitude lower complexity. 

The rest of the paper is organized as follows: In Section \ref{sec:relatedwork} we summarize the state-of-the-art works, especially in video classification using Neural Network models and the tensorization of weight matrices. 
In Section \ref{sec:ttrnn} we first introduce the Tensor-Train model and then provide a detailed derivation of our proposed Tensor-Train RNNs.
In Section \ref{sec:experiments} we present our experimental results on three large scale video datasets. Finally, Section \ref{sec:conclusion} serves as a wrap-up of our current contribution and provides an outlook of future work. 

\paragraph{Notation} 
We index an entry in a $d$-dimensional tensor $\bs{\mathcal{A}} \in \mathbb{R}^{p_1 \times p_2 \times ... \times p_d}$ using round parentheses such as $\bs{\mathcal{A}}(l_1, l_2, ..., l_d) \in \mathbb{R}$ and $\bs{\mathcal{A}}(l_1) \in \mathbb{R}^{p_2 \times p_3 \times ... \times p_d}$, when we only write the first index. Similarly, we also use $\bs{\mathcal{A}}(l_1, l_2) \in \mathbb{R}^{p_3 \times p_4 \times ... \times p_d}$ to refer to the sub-tensor specified by two indices $l_1$ and $l_2$.

\section{Related Works} \label{sec:relatedwork}

The current approaches to model video data are closely related to models for image data. 
A large majority of these works use deep CNNs to process each frame as image, and aggregate the CNN outputs.
\cite{karpathy2014large} proposes multiple fusion techniques such as Early, Late and Slow Fusions, covering different aspects of the video. This approach, however, does not fully take the order of frames into account.
\cite{yue2015beyond} and \cite{fernando2016learning} apply global pooling of frame-wise CNNs, before feeding the aggregated information to the final classifier.
An intuitive and appealing idea is to fuse these frame-wise spatial representations learned by CNNs using RNNs. The major challenge, however, is the computation complexity; and for this reason multiple compromises in the model design have to be made: \cite{sr2015uns} restricts the length of the sequences to be 16, while \cite{sharma2015action} and \cite{donahue2015long} use pre-trained CNNs. 
\cite{xingjian2015convolutional} proposed a more compact solution that applies convolutional layers as input-to-hidden and hidden-to-hidden mapping in LSTM. However, they did not show its performance on large-scale video data. 
\cite{simonyan2014two} applied two stacked CNNs, one for spatial features and the other for temporal ones, and fused the outcomes of both using averaging and a Support-Vector Machine as classifier. This approach is further enhanced with Residual Networks in \cite{fe2016spa}. 
To the best of our knowledge, there has been no published work on applying pure RNN models to video classification or related tasks.

The Tensor-Train was first introduced by \cite{oseledets2011tensor} as a tensor factorization model with the advantage of being capable of scaling to an arbitrary number of dimensions. \cite{novikov2015tensorizing} showed that one could reshape a fully connected layer into a high-dimensional tensor and then factorize this tensor using Tensor-Train. This was applied to compress very large weight matrices in deep Neural Networks where the entire model was trained end-to-end. In these experiments they compressed fully connected layers on top of convolution layers, and also proved that a Tensor-Train Layer can directly consume pixels of image data such as CIFAR-10, achieving the best result among all known non-convolutional models. Then in \cite{garipov2016ultimate} it was shown that even the convolutional layers themselves can be compressed with Tensor-Train Layers. 
Actually, in an earlier work by \cite{lebedev2014speeding} a similar approach had also been introduced, but their CP factorization is calculated in a pre-processing step and is only fine tuned with error back propagation as a post processing step.

\cite{koutnik2014clockwork} performed two sequence classification tasks using multiple RNN architectures of relatively low dimensionality: The first task was to classify spoken words where the input sequence had a dimension of 13 channels. In the second task, RNNs were trained to classify handwriting based on the time-stamped 4D spatial features. RNNs have been also applied to classify the sentiment of a sentence such as in the IMDB reviews dataset \cite{maas2011learning}. In this case, the word embeddings form the input to RNN models and they may have a dimension of a few hundreds. The sequence classification model can be seen as a special case of the Encoder-Decoder-Framework \cite{sutskever2014sequence} in the sense that a classifier decodes the learned representation for the entire sequence into a probabilistic distribution over all classes.

\section{Tensor-Train RNN} \label{sec:ttrnn}
In this section, we first give an introduction to the core ingredient of our proposed approach, i.e., the Tensor-Train Factorization, and then use this to formulate a so-called Tensor-Train Layer \cite{novikov2015tensorizing} which replaces the weight matrix mapping from the input vector to the hidden layer in RNN models. We emphasize that such a layer is learned end-to-end, together with the rest of the RNN in a very efficient way. 
\subsection{Tensor-Train Factorization}
\begin{figure}[h]
\begin{center}
	\includegraphics[scale=.26, clip=true, trim=13cm 5cm 8cm 7.5cm]{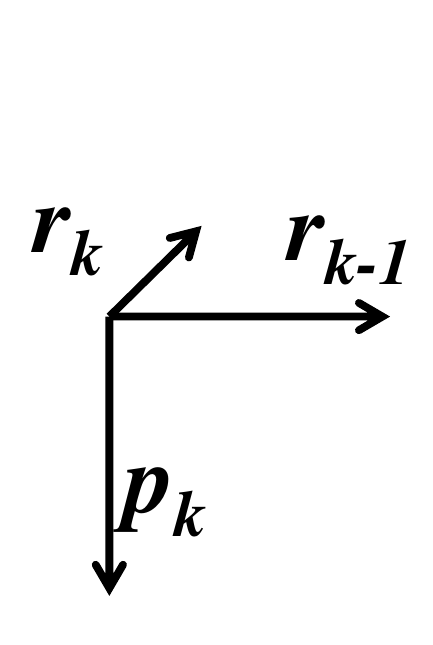}
	\includegraphics[scale=.26, clip=true, trim=0cm 5cm 0cm 7.5cm]{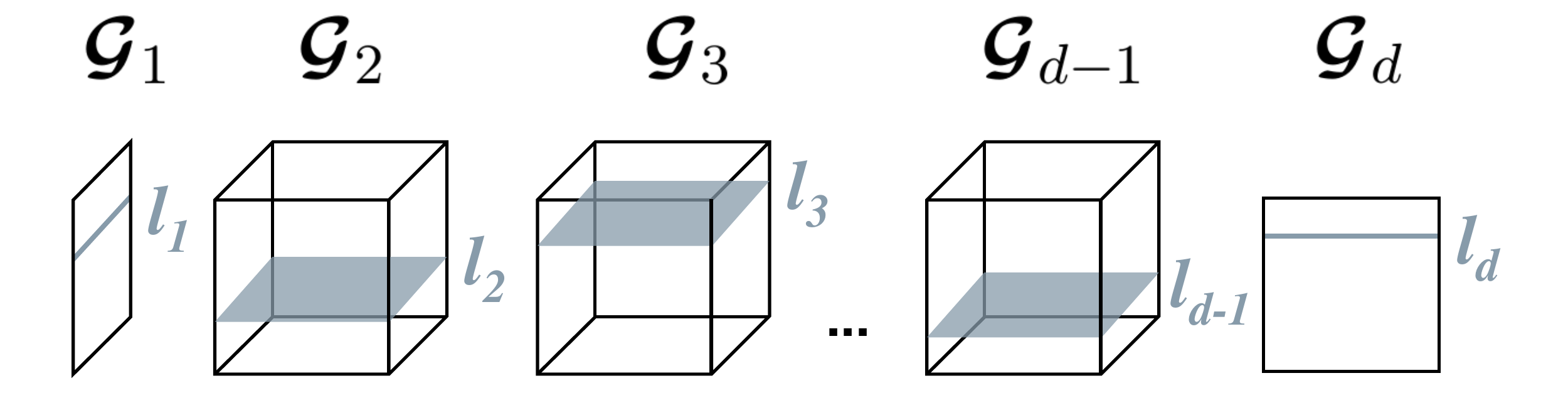}
	\caption{Tensor-Train Factorization Model: To reconstruct one entry in the target tensor, one performs a sequence of vector-matrix-vector multiplications, yielding a scalar. }
	\label{fig:TTF}
\end{center}
\end{figure}

A \emph{Tensor-Train Factorization} (TTF) is a tensor factorization model that can scale to an arbitrary number of dimensions. Assuming a $d$-dimensional target tensor of the form $ \bs{\mathcal{A}} \in \mathbb{R} ^{p_1 \times p_2 \times ... \times p_d} $, it can be factorized in form of:
\begin{align} \label{eq:TTF_1}
	\skew{6}{\widehat}{\bs{\mathcal{A}}}(l_1, l_2, ..., l_d) &\stackrel{TTF}{=} \bs{\mathcal{G}}_1(l_1) ~ \bs{\mathcal{G}}_2(l_2) ~ ... ~ \bs{\mathcal{G}}_d(l_d)
\end{align}
where
\begin{align} \label{eq:TTF_2}
	\begin{split}
	&\bs{\mathcal{G}}_k \in \mathbb{R}^{p_k \times r_{k-1} \times r_k}, ~l_k  \in [1, p_k] ~\forall k \in [1, d] \\
	&\text{and} ~~ r_0 = r_d = 1.
	\end{split}
\end{align}
As Eq. \ref{eq:TTF_1} suggests, each entry in the target tensor is represented as a sequence of matrix multiplications. The set of tensors $\{ \bs{\mathcal{G}}_k \}_{k=1}^{d}$ are usually called core-tensors. The complexity of the TTF is determined by the ranks $[r_0, r_1, ..., r_d]$. We demonstrate this calculation also in Fig. \ref{fig:TTF}. 
Please note that the dimensions and core-tensors are indexed from $1$ to $d$ while the rank index starts from 0; also note that the first and last ranks are both restricted to be $1$, which implies that the first and last core tensors can be seen as matrices so that the outcome of the chain of multiplications in Eq. \ref{eq:TTF_1} is always a scalar.

If one imposes the constraint that each integer $p_k$ as in Eq. \eqref{eq:TTF_1} can be factorized as $p_k = m_k \cdot n_k ~\forall k \in [1, d]$, and consequently reshapes each $\bs{\mathcal{G}}_k$ into $\bs{\mathcal{G}}_k^{*} \in \mathbb{R}^{m_k \times n_k \times r_{k-1} \times r_k}$, then each index $l_k$ in Eq. \eqref{eq:TTF_1} and \eqref{eq:TTF_2} can be uniquely represented with two indices $(i_k, j_k)$, i.e.
\begin{align} \label{eq:TTL_1}
&i_k = \lfloor \frac{l_k}{n_k} \rfloor, ~ j_k = l_k - n_k \lfloor \frac{l_k}{n_k} \rfloor, \\
&\text{so that }~ \bs{\mathcal{G}}_k(l_k) = \bs{\mathcal{G}}^{*}_k(i_k, j_k) \in  \mathbb{R}^{r_{k-1} \times r_k}.
\end{align}
Correspondingly, the factorization for the tensor $\bs{\mathcal{A}} \in \mathbb{R}^{(m_1 \cdot n_1) \times (m_2 \cdot n_2) \times ... \times (m_d \cdot n_d)}$ can be rewritten  equivalently to Eq.\eqref{eq:TTF_1}:
\begin{align} \label{eq:TTL_2}
\begin{split}
	& \skew{6}{\widehat}{\bs{\mathcal{A}}}((i_1, j_1), (i_2, j_2), ..., (i_d, j_d)) \\
	\stackrel{TTF}{=} & \bs{\mathcal{G}}_1^{*}(i_1, j_1) ~ \bs{\mathcal{G}}_2^{*}(i_2, j_2) ~ ... ~ \bs{\mathcal{G}}_d^{*}(i_d, j_d). 
\end{split}
\end{align}
This double index trick \cite{novikov2015tensorizing} enables the factorizing of weight matrices in a feed-forward layer as described next.

\subsection{Tensor-Train Factorization of a Feed-Forward Layer}
Here we factorize the weight matrix $\bs{W}$ of a fully-connected feed-forward layer denoted in $\hat{\bs{y}} = \bs{W} \bs{x} +\bs{b}$.

First we rewrite this layer in an equivalent way with scalars as:
\begin{align} \label{eq:TTL_3}
	\begin{split}
	& \hat{\bs{y}}(j) = \sum_{i=1}^{M} \bs{W}(i,j) \cdot \bs{x}(i) +\bs{b}(j) \\
	&\forall j \in [1, N] \text{ and with}~~ \bs{x} \in \mathbb{R}^{M}, ~\bs{y} \in \mathbb{R}^{N}.
	\end{split}
\end{align}
Then, if we assume that $M = \prod_{k=1}^{d}m_k, ~~ N = \prod_{k=1}^{d}n_k $ i.e. both $M$ and $N$ can be factorized into two integer arrays of the same length, then we can reshape the input vector $\bs{x}$ and the output vector $\hat{\bs{y}}$ into two tensors with the same number of dimensions: $\bs{\mathcal{X}} \in \mathbb{R}^{m_1 \times m_2 \times ... \times m_d}, \bs{\mathcal{Y}} \in \mathbb{R}^{n_1 \times n_2 \times ... \times n_d}$,  and the mapping function $\mathbb{R}^{m_1 \times m_2 \times ... \times m_d} \rightarrow \mathbb{R}^{n_1 \times n_2 \times ... \times n_d}$ can be written as:
\begin{align}
	\begin{split} \label{eq:TTL_4}
	& \widehat{\bs{\mathcal{Y}}}(j_1, j_2, ..., j_d) \\
	& ~~= \sum_{i_1=1}^{m_1} \sum_{i_2=1}^{m_2} ... \sum_{i_d=1}^{m_d} \bs{\mathcal{W}}((i_1, j_1), (i_2, j_2), ..., (i_d, j_d)) \cdot \\
	& ~~~~~~\bs{\mathcal{X}}(i_1, i_2, ..., i_d) + \bs{\mathcal{B}}(j_1, j_2, ..., j_d). \\
	\end{split}
\end{align}

Note that Eq. \eqref{eq:TTL_3} can be seen as a special case of Eq. \eqref{eq:TTL_4} with $d=1$.
The $d$-dimensional double-indexed tensor of weights $\bs{\mathcal{W}}$ in Eq.\eqref{eq:TTL_4} can be replaced by its TTF representation:
\begin{align}
	\begin{split} \label{eq:TTL_5}
	& \widehat{\bs{\mathcal{W}}}((i_1, j_1), (i_2, j_2), ..., (i_d, j_d)) \\
	& \stackrel{TTF}{=} \bs{\mathcal{G}}_1^{*}(i_1, j_1) ~ \bs{\mathcal{G}}_2^{*}(i_2, j_2) ~ ... ~ \bs{\mathcal{G}}_d^{*}(i_d, j_d).
	\end{split}
\end{align}
Now instead of explicitly storing the full tensor $\bs{\mathcal{W}}$ of size $\prod_{k=1}^{d}m_k \cdot n_k = M \cdot N$, we only store its TT-format, i.e., the set of low-rank core tensors $\{ \bs{\mathcal{G}}_k \}_{k=1}^{d}$ of size $\sum_{k=1}^{d} m_k \cdot n_k \cdot r_{k-1} \cdot r_k$, which can approximately reconstruct $\bs{\mathcal{W}}$.

The forward pass complexity \cite{novikov2015tensorizing} for one scalar in the output vector indexed by $(j_1, j_2, ..., j_d)$ turns out to be $\mathcal{O}(d \cdot \tilde{m} \cdot \tilde{r}^2)$. Since one needs an iteration through all such tuples, yielding $\mathcal{O}(\tilde{n}^d)$, the total complexity for one Feed-Forward-Pass can be expressed as $\mathcal{O}(d \cdot \tilde{m} \cdot \tilde{r}^2 \cdot \tilde{n}^d )$, where $ \tilde{m} = \max_{k\in[1,d]}m_k, \tilde{n} = \max_{k\in[1,d]}n_k, \tilde{r} = \max_{k\in[1,d]}r_k$. This, however, would be $\mathcal{O}(M \cdot N)$ for a fully-connected layer.

One could also compute the compression rate as the ratio between the number of weights in a fully connected layer and that in its compressed form as:
\begin{align}  \label{eq:TTL_6}
	r = \frac{\sum_{k=1}^{d} m_k n_k r_{k-1} r_k}{\prod_{k=1}^{d} m_k n_k}. 
\end{align}

For instance, an RGB frame of size 160 $\times$ 120 $\times$ 3 implies an input vector of length 57,600. With a hidden layer of size, say, 256 one would need a weight matrix consisting of 14,745,600 free parameters. On the other hand, a TTL that factorizes the input dimension with 8$\times$20$\times$20$\times$18 is able to represent this matrix using 2,976 parameters with a TT-rank of 4, or 4,520 parameters with a TT-rank of 5 (Tab. \ref{tab:numexample}), yielding compression rates of 2.0\textit{e}-4 and 3.1\textit{e}-4, respectively.

For the rest of the paper, we term a fully-connected layer in form of $\hat{\bs{y}} = \bs{W} \bs{x} + \bs{b}$, whose weight matrix $\bs{W}$ is factorized with TTF, a \emph{Tensor-Train Layer} (TTL) and use the notation 
\begin{align}
	\hat{\bs{y}} = TTL( \bs{W}, \bs{b}, \bs{x}), \text{or}~~  TTL( \bs{W}, \bs{x})
\end{align}
where in the second case no bias is required. Please also note that, in contrast to \cite{lebedev2014speeding} where the weight tensor is firstly factorized using non-linear Least-Square method and then fine-tuned with Back-Propagation, the TTL is always trained end-to-end. For details on the gradients calculations please refer to Section 5 in \cite{novikov2015tensorizing}.

\subsection{Tensor-Train RNN}
In this work we investigate the challenge of modeling high-dimensional sequential data with RNNs. For this reason, we factorize the matrix mapping from the input to the hidden layer with a TTL. 
For an Simple RNN (SRNN), which is also known as the Elman Network, this mapping is realized as a vector-matrix multiplication, whilst in case of LSTM and GRU, we consider the matrices that map from the input vector to the gating units:  

TT-GRU:
\begin{align}	\label{eq:tt_gru}
	\begin{split}
	\bs{r}^{[t]} &= \sigma (TTL(\bs{W}^r, \bs{x}^{[t]}) + \bs{U}^r \bs{h}^{[t-1]} + \bs{b}^r) \\
	\bs{z}^{[t]} &= \sigma (TTL(\bs{W}^z, \bs{x}^{[t]}) + \bs{U}^z \bs{h}^{[t-1]} + \bs{b}^z) \\	
	\bs{d}^{[t]} &= \tanh (TTL(\bs{W}^d, \bs{x}^{[t]}) + \bs{U}^d (\bs{r}^{[t]} \circ \bs{h}^{[t-1]}) ) \\
	\bs{h}^{[t]} &= (1 - \bs{z}^{[t]}) \circ \bs{h}^{[t-1]} + \bs{z}^{[t]} \circ \bs{d}^{[t]}, 
	\end{split}	
\end{align}

TT-LSTM:
\begin{align}	\label{eq:tt_lstm}
	\begin{split}
	\bs{k}^{[t]} &= \sigma ( TTL(\bs{W}^k, \bs{x}^{[t]}) + \bs{U}^k \bs{h}^{[t-1]} + \bs{b}^k) \\
	\bs{f}^{[t]} &= \sigma ( TTL(\bs{W}^f, \bs{x}^{[t]}) + \bs{U}^f \bs{h}^{[t-1]} + \bs{b}^f) \\
	\bs{o}^{[t]} &= \sigma ( TTL(\bs{W}^o, \bs{x}^{[t]}) + \bs{U}^o \bs{h}^{[t-1]} + \bs{b}^o) \\
	\bs{g}^{[t]} &= \tanh ( TTL(\bs{W}^g, \bs{x}^{[t]}) + \bs{U}^g \bs{h}^{[t-1]} + \bs{b}^g) \\
	\bs{c}^{[t]} &= \bs{f}^{[t]} \circ \bs{c}^{[t-1]} + \bs{k}^{[t]} \circ \bs{g}^{[t]}\\
	\bs{h}^{[t]} &= \bs{o}^{[t]} \circ \tanh(\bs{c}^{[t]}). \\
	\end{split}	
\end{align}

One can see that LSTM and GRU require 4 and 3 TTLs, respectively, one for each of the gating units. Instead of calculating these TTLs successively (which we call vanilla TT-LSTM and vanilla TT-GRU), we increase $n_1$ ---the first \footnote{Though in theory one could of course choose any $n_k$.} of the factors that form the output size $N=\prod_{k=1}^{d}n_k$ in a TTL--- by a factor of 4 or 3, and concatenate all the gates as one output tensor, thus parallelizing the computation. 
This trick, inspired by the implementation of standard LSTM and GRU in \cite{keras2015}, can further reduce the number of parameters, where the concatenation is actually participating in the tensorization. 
The compression rate for the input-to-hidden weight matrix $\bs{W}$ now becomes 
\begin{align}
	& r^{*} = \frac{\sum_{k=1}^{d} m_k n_k r_{k-1} r_k + (c-1) (m_1 n_1 r_0 r_1)}{c \cdot \prod_{k=1}^{d} m_k n_k} \\
	& \text{where}~ c=4 ~\text{in case of LSTM and }~ 3 ~\text{in case of GRU},  \nonumber
\end{align}
and one can show that $r^{*}$ is always smaller than $r$ as in Eq. \ref{eq:TTL_6}. For the former numerical example of a input frame size 160$\times$120$\times$3, a vanilla TT-LSTM would simply require 4 times as many parameters as a TTL, which would be 11,904 for rank 4 and 18,080 for rank 5. Applying this trick would, however, yield only 3,360 and 5,000 parameters for both ranks, respectively. We cover other possible settings of this numerical example in Tab. \ref{tab:numexample}. 

\begin{table*}[t]
\caption{A numerical example of compressing with TT-RNNs. Assuming that an input dimension of 160$\times$120$\times$3 is factorized as 8 $\times$ 20 $\times$ 20 $\times$ 18 and 
the hidden layer as $4 \times 4 \times 4 \times 4 =256$, depending on the TT-ranks we calculate the number of parameters necessary for a Fully-Connected (FC) layer, a TTL which is equivalent to TT-SRNN , TT-LSTM and TT-GRU in their respective vanilla and parallelized form. For comparison, typical CNNs for preprocessing images such as AlexNet \cite{krizhevsky2012imagenet, han2015learning} or GoogLeNet \cite{szegedy2015going} consist of over 61 and 6 million parameters, respectively. }
\label{tab:numexample}
\begin{center}
\begin{small}
\begin{tabular}{ | c | r r r r r r|}  \hline
FC 								& TT-ranks & TTL & vanilla TT-LSTM	& TT-LSTM 	& vanilla TT-GRU	& TT-GRU \\ \hline \hline
\multirow{3}{*}{14,745,600}		& 3				& 1,752			& 7,008				& 2,040		& 5,256				& 1,944 \\
									& 4				& 2,976			& 11,904				& 3,360		& 8,928				& 3,232 \\
									& 5				& 4,520			& 18,080				& 5,000		& 13,560				& 4,840 \\ \hline
\end{tabular}
\end{small}
\end{center}
\end{table*}


Finally to construct the classification model, we denote the $i$-th sequence of variable length $T_i$ as a set of vectors $\{ \bs{x}_i^{[t]} \}_{t=1}^{T_i}$ with $\bs{x}_i^{[t]} \in \mathbb{R}^{M} \forall t $. For video data each $\bs{x}_i^{[t]}$ would be an RGB frame of 3 dimensions. For the sake of simplicity we denote an RNN model, either with or without TTL, with a function $f(\cdot)$:
\begin{align}
	\bs{h}_i^{[T_i]} = f(\{ \bs{x}_i^{[t]} \}_{t=1}^{T_i}), ~\text{where}~ \bs{h}_i^{[T_i]} \in{\mathbb{R}^{N}},
\end{align}
which outputs the last hidden layer vector $\bs{h}_i^{[T_i]}$ out of a sequential input of variable length. This vector can be interpreted as a latent representation of the whole sequence, on top of which a parameterized classifier $\phi(\cdot)$ with either softmax or logistic activation produces the distribution over all J classes:
\begin{align}
	\begin{split}
	\mathbb{P}(\bs{y}_{i} = \bs{1} | \{ \bs{x}_i^{[t]} \}_{t=1}^{T_i}) 	&= \phi (\bs{h}_i^{[T_i]}) \\
																							&= \phi (f(\bs{x}_i^{[t]} \}_{t=1}^{T_i})) \in{[0,1]^{J}},
	\end{split}
\end{align}
The model is also illustrated in Fig. \ref{fig:model_architecture}:
\begin{figure*}[t]
\begin{center}
\includegraphics[scale=.6, clip=true, trim=0cm 5.2cm 0cm 5cm]{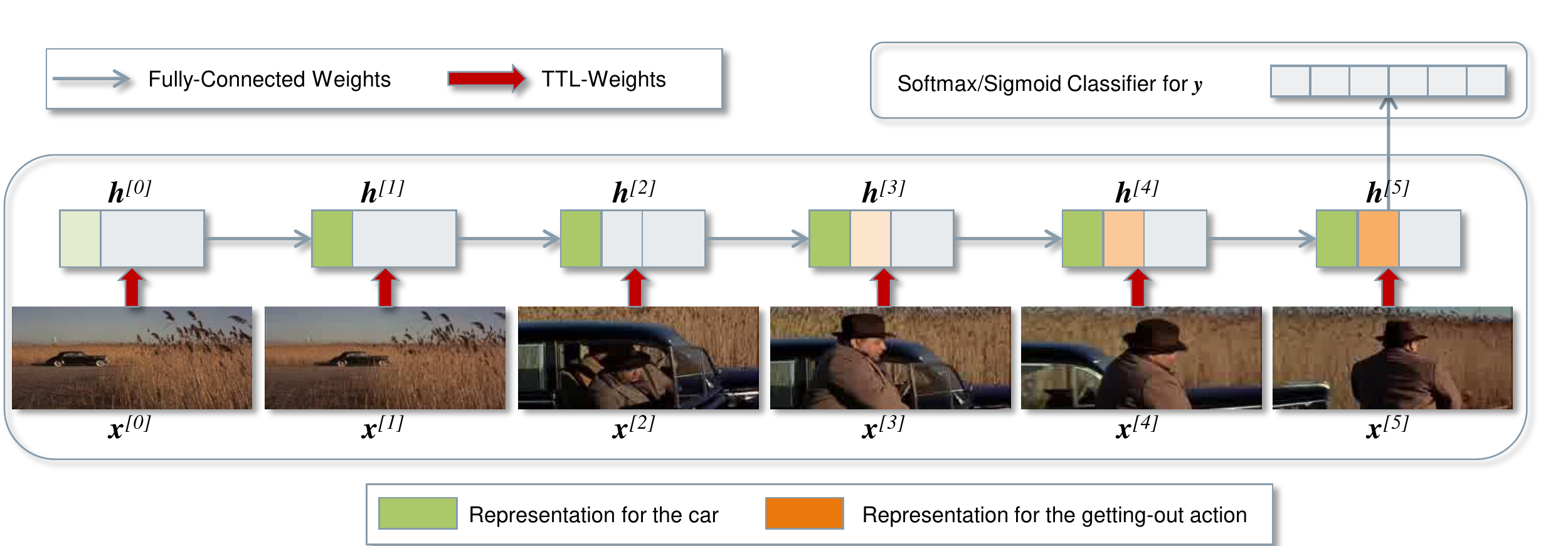}
\caption{Architecture of the proposed model based on TT-RNN (For illustrative purposes we only show 6 frames): A softmax or sigmoid classifier built on the last hidden layer of a TT-RNN. We hypothesize that the RNN can be encouraged to aggregate the representations of different shots together and produce a global representation for the whole sequence.}
\label{fig:model_architecture}
\end{center}
\end{figure*}

\section{Experiments} \label{sec:experiments}
In the following, we present our experiments conducted on three large video datasets.
These empirical results demonstrate that the integration of the Tensor-Train Layer in plain RNN architectures such as a tensorized LSTM or GRU boosts the classification quality of these models tremendously when directly exposed to high-dimensional input data, such as video data.
In addition, even though the plain architectures are of very simple nature and very low complexity opposed to the state-of-the-art solutions on these datasets, it turns out that the integration of the Tensor-Train Layer alone makes these simple networks very competitive to the state-of-the-art, reaching second best results in all cases. 

\paragraph{UCF11 Data \cite{liu2009recognizing}} \mbox{}\\
We first conduct experiments on the UCF11 -- earlier known as the YouTube Action Dataset. It contains in total 1600 video clips belonging to 11 classes that summarize the human action visible in each video clip such as basketball shooting, biking, diving etc.. These videos originate from YouTube and have natural background ('in the wild') and a resolution of 320 $\times$ 240. We generate a sequence of RGB frames of size 160 $\times$ 120 from each clip at an fps(frame per second) of 24, corresponding to the standard value in film and television production. The lengths of frame sequences vary therefore between 204 to 1492 with an average of 483.7.

\begin{figure}[h]
\begin{center}
\includegraphics[width=.2\columnwidth]{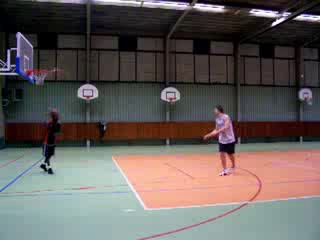}
\includegraphics[width=.2\columnwidth]{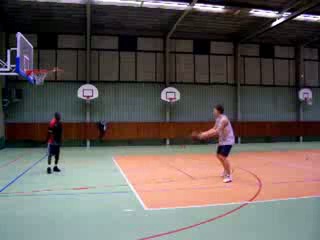}
\includegraphics[width=.2\columnwidth]{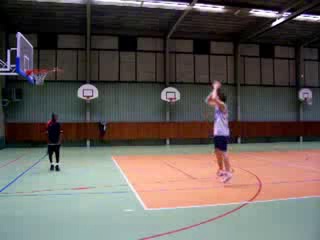}
\includegraphics[width=.2\columnwidth]{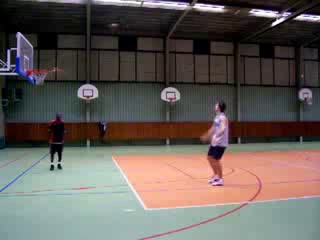} \\

\includegraphics[width=.2\columnwidth]{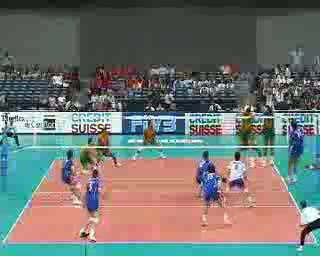}
\includegraphics[width=.2\columnwidth]{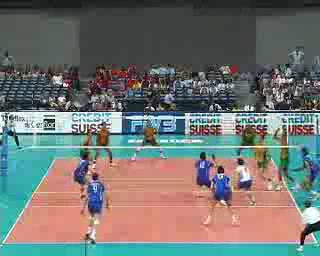}
\includegraphics[width=.2\columnwidth]{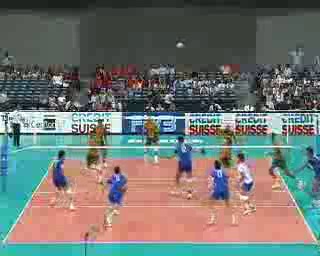}
\includegraphics[width=.2\columnwidth]{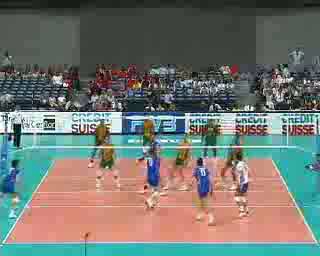} \\

\caption{Two samples of frame sequences from the UCF11 dataset. The two rows belong to the classes of basketball shooting and volleyball spiking, respectively. }
\label{fig:ucf11_snaps}
\end{center}
\end{figure}

For both the TT-GRUs and TT-LSTMs the input dimension at each time step is $160 \times 120 \times 3 = 57600$ which is factorized as $8 \times 20 \times 20 \times 18$, the hidden layer is chosen to be $4 \times 4 \times 4 \times 4 = 256$ and the Tensor-Train ranks are $[1, 4, 4, 4, 1]$. A fully-connected layer for such a mapping would have required 14,745,600 parameters to learn, while the input-to-hidden layer in TT-GRU and TT-LSTM consist of only 3,360 and 3,232, respectively. 

As the first baseline model we sample 6 random frames in ascending order. The model is a simple Multilayer Perceptron (MLP) with two layers of weight matrices, the first of which being a TTL. The input is the concatenation of all 6 flattened frames and the hidden layer is of the same size as the hidden layer in TT-RNNs. We term this model as Tensor-Train Multilayer Perceptron (TT-MLP) for the rest of the paper. 
As the second baseline model we use plain GRUs and LSTMs that have the same size of hidden layer as their TT pendants.
We follow \cite{liu2013spatial} and perform for each experimental setting a 5-fold cross validation with mutual exclusive data splits. The mean and standard deviation of the prediction accuracy scores are reported in Tab. \ref{tab:ucf11_res}.


\begin{table}[h]
\caption{Experimental Results on UCF11 Dataset. We report i) the accuracy score, ii) the number of parameters involved in the input-to-hidden mapping in respective models and iii) the average runtime of each training epoch. The models were trained on a Quad core Intel\textregistered Xeon\textregistered E7-4850 v2 2.30GHz Processor to a maximum of 100 epochs}
\label{tab:ucf11_res}
\begin{center}
\begin{small}
\begin{tabular}{ | l  c  r  r |}  \hline
					& Accuracy 				& \# Parameters & Runtime\\ \hline\hline
TT-MLP			& 0.427 $\pm$ 0.045	&  7,680				& 902s\\
GRU				& 0.488 $\pm$ 0.033& 	44,236,800 		& 7,056s\\
LSTM				& 0.492 $\pm$ 0.026 &	58,982,400 		& 8,892s\\
TT-GRU			& \textbf{0.813} $\pm$ \textbf{0.011} 	& 3,232 	& 1,872s\\
TT-LSTM			& 0.796 $\pm$ 0.035 	& 3,360 			& 2,160s \\ \hline 
\end{tabular}
\end{small}
\end{center}
\end{table}

The standard LSTM and GRU do not show large improvements compared with the TT-MLP model. The TT-LSTM and TT-GRU, however, do not only compress the weight matrix from over 40 millions to 3 thousands, but also significantly improve the classification accuracy. It seems that plain LSTM and GRU are not adequate to model such high-dimensional sequential data because of the large weight matrix from input to hidden layer. Compared to some latest state-of-the-art performances in Tab. \ref{tab:ucf11_other}, our model ---simple as it is--- shows accuracy scores second to \cite{sharma2015action}, which uses pre-trained GoogLeNet CNNs plus 3-fold stacked LSTM with attention mechanism. Please note that a GoogLeNet CNN alone consists of over 6 million parameters \cite{szegedy2015going}. 
In term of runtime, the plain GRU and LSTM took on average more than 8 and 10 days to train, respectively; while the TT-GRU and TT-LSTM both approximately 2 days. Therefore please note the TTL reduces the training time by a factor of 4 to 5 on these commodity hardwares. 

\begin{table}[h]
\caption{State-of-the-art results on the UCF11 Dataset, in comparison with our best model. Please note that there was an update of the data set on 31th December 2011. We therefore only consider works posterior to this date.}
\label{tab:ucf11_other}
\begin{center}
\begin{small}
\begin{tabular}{| l  | c | } \hline
Original: \cite{liu2009recognizing} 		& 0.712  	\\ \hline \hline
\cite{liu2013spatial}				& 0.761	\\
\cite{hasan2014incremental}	& 0.690	\\
\cite{sharma2015action}			& \textbf{0.850} \\ \hline \hline
Our best model (TT-GRU)		& 0.813 \\
\hline
\end{tabular}
\end{small}
\end{center}
\vskip -0.1in
\end{table}
 
\paragraph{Hollywood2 Data \cite{marszalek09}}\mbox{}\\
The Hollywood2 dataset contains video clips from 69 movies, from which 33 movies serve as training set and 36 movies as test set. From these movies 823 training clips and 884 test clips are generated and each clip is assigned one or multiple of 12 action labels such as answering the phone, driving a car, eating or fighting a person. This data set is much more realistic and challenging since the same action could be performed in totally different style in front of different background in different movies. Furthermore, there are often montages, camera movements and zooming within a single clip.

\begin{figure}[h]
\begin{center}
\includegraphics[width=.3\columnwidth]{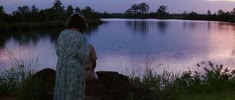}
\includegraphics[width=.3\columnwidth]{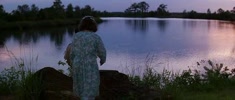}
\includegraphics[width=.3\columnwidth]{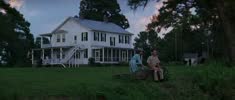}\\
\includegraphics[width=.3\columnwidth]{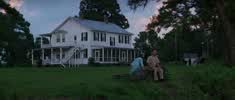}
\includegraphics[width=.3\columnwidth]{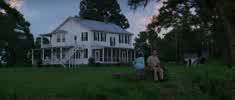}
\includegraphics[width=.3\columnwidth]{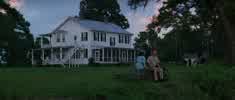} \\
\includegraphics[width=.3\columnwidth]{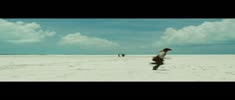}
\includegraphics[width=.3\columnwidth]{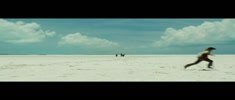}
\includegraphics[width=.3\columnwidth]{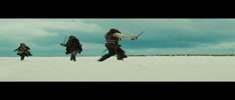}\\
\includegraphics[width=.3\columnwidth]{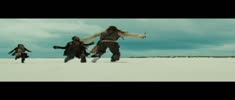}
\includegraphics[width=.3\columnwidth]{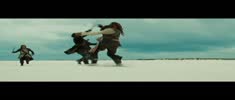}
\includegraphics[width=.3\columnwidth]{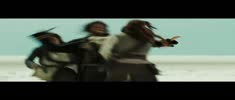} \\

\caption{Two samples of frame sequences from the Hollywood2 dataset. The first sequence (row 1 and 2) belongs to the class of sitting down; the second sequence (row 3 and 4) has two labels: running and fighting person. }
\label{fig:HW2_snaps}
\end{center}
\end{figure}

The original frame sizes of the videos vary, but based on the majority of the clips we generate frames of size 234 $\times$ 100, which corresponds to the Anamorphic Format, at fps of 12.
The length of training sequences varies from 29 to 1079 with an average of 134.8; while the length of test sequences varies from 30 to 1496 frames with an average of 143.3.

The input dimension at each time step, being $234 \times 100 \times 3 = 70200$, is factorized as $10 \times 18 \times 13 \times 30$. The hidden layer is still $4 \times 4 \times 4 \times 4 = 256$ and the Tensor-Train ranks are $[1, 4, 4, 4, 1]$. Since each clip might have more than one label (multi-class multi-label problem) we implement a logistic activated classifier for each class on top of the last hidden layer. Following \cite{marszalek09} we measure the performances using Mean Average Precision across all classes, which corresponds to the Area-Under-Precision-Recall-Curve.

As before we conduct experiments on this dataset using the plain LSTM, GRU and their respective TT modifications. The results are presented in in Tab. \ref{tab:hw2_res} and state-of-the-art in Tab. \ref{tab:hw2_other}. 

\begin{table}[h]
\caption{Experimental Results on Hollywood2 Dataset. We report i) the Mean Average Precision score, ii) the number of parameters involved in the input-to-hidden mapping in respective models and iii) the average runtime of each training epoch. The models were trained on an NVIDIA Tesla K40c Processor to a maximum of 500 epochs. }
\label{tab:hw2_res}
\begin{center}
\begin{small}
\begin{tabular}{ | l  c  r  r |}  \hline
					& MAP	 					& \# Parameters & Runtime \\ \hline\hline
TT-MLP				& 0.103	& 4,352 & 16s \\
GRU				& 0.249 	& 53,913,600 	& 106s \\
LSTM				& 0.108 	& 71,884,800 	& 179s \\
TT-GRU			& 0.537	& 2,944 			& 96s  \\
TT-LSTM			& \textbf{0.546}	& 3,104	& 102s \\ \hline
\end{tabular}
\end{small}
\end{center}
\end{table}

\cite{fernando2015modeling} and \cite{jain2013better} use improved trajectory features with Fisher encoding \cite{wang2013action} and Histogram of Optical Flow (HOF) features \cite{laptev2008learning}, respectively, and achieve so far the best score. \cite{sharma2015action} and \cite{fernando2016learning} provide best scores achieved with Neural Network models but only the latter applies end-to-end training. To this end, the TT-LSTM model provides the second best score in general and the best score with Neural Network models, even though it merely replaces the input-to-hidden mapping with a TTL. Please note the large difference between the plain LSTM/GRU and the TT-LSTM/GRU, which highlights the significant  performance improvements the Tensor-Train Layer contributes to the RNN models. 

It is also to note that, although the plain LSTM and GRU consist of up to approximately 23K as many parameters as their TT modifications do, the training \emph{time} does not reflect such discrepancy due to the good parallelization power of GPUs. However, the obvious difference in their training \emph{qualities} confirms that training larger models may require larger amounts of data. 
In such cases, powerful hardwares are no guarantee for successful training. 

\begin{table}[h]
\caption{State-of-the-art Results on Hollywood2 Dataset, in comparison with our best model. }
\label{tab:hw2_other}
\begin{center}
\begin{small}
\begin{tabular}{| l  | c | } \hline
Original: \cite{marszalek09} 			& 0.326 \\ \hline \hline
\cite{le2011learning} 		& 0.533 \\
\cite{jain2013better}			& 0.542 \\
\cite{sharma2015action}		& 0.439 \\
\cite{fernando2015modeling} & \textbf{0.720} \\
\cite{fernando2016learning} 		& 0.406 \\ \hline \hline
Our best model (TT-LSTM)			& 0.546 \\ \hline
\end{tabular}
\end{small}
\end{center}
\end{table}

\paragraph{Youtube Celebrities Face Data \cite{kim2008face}} \mbox{} \\
This dataset consists of 1910 Youtube video clips of 47 prominent individuals such as movie stars and politicians. In the simplest cases, where the face of the subject is visible as a long take, a mere frame level classification would suffice. The major challenge, however, is posed by the fact that some videos involve zooming and/or changing the angle of view. In such cases a single frame may not provide enough information for the classification task and we believe it is advantageous to apply RNN models that can aggregate frame level information over time. 

\begin{figure}[h]
\begin{center}
\includegraphics[width=.2\columnwidth]{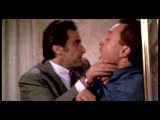}
\includegraphics[width=.2\columnwidth]{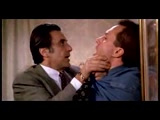}
\includegraphics[width=.2\columnwidth]{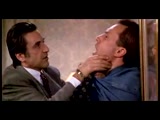}
\includegraphics[width=.2\columnwidth]{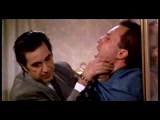}
\includegraphics[width=.2\columnwidth]{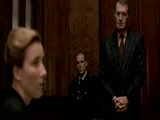}
\includegraphics[width=.2\columnwidth]{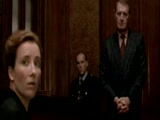}
\includegraphics[width=.2\columnwidth]{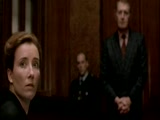}
\includegraphics[width=.2\columnwidth]{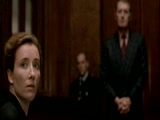}

\caption{Two samples of frame sequences from the Youtube Celebrities Face dataset. The two rows belong to the classes of Al Pacino and Emma Thompson.}
\label{fig:YCF_snaps}
\end{center}
\end{figure}

The original frame sizes of the videos vary but based on the majority of the clips we generate frames of size 160 $\times$ 120 at fps of 12. The retrieved sequences have lengths varying from 2 to 85 with an average of 39.9.
The input dimension at each time step is $160 \times 120 \times 3 = 57600$ which is factorized as $4 \times 20 \times 20 \times 36$, the hidden layer is again $4 \times 4 \times 4 \times 4 = 256$ and the Tensor-Train ranks are $[1, 4, 4, 4, 1]$.

\begin{table}[h]
\caption{Experimental Results on Youtube Celebrities Face Dataset. We report i) the Accuracy score, ii) the number of parameters involved in the input-to-hidden mapping in respective models and iii) the average runtime of each training epoch. The models were trained on an NVIDIA Tesla K40c Processor to a maximum of 100 epochs. }
\label{tab:ycf_res}
\vskip -0.2in
\begin{center}
\begin{small}
\begin{tabular}{ | l  c  r  r |}  \hline
					& Accuracy 					& \# Parameters & Runtime \\ \hline\hline
TT-MLP				& 0.512 $\pm$ 0.057		& 3,520 	& 14s\\
GRU				& 0.342 $\pm$ 0.023		& 38,880,000 & 212s\\
LSTM				& 0.332	$\pm$ 0.033	& 51,840,000 & 253s\\
TT-GRU			& \textbf{0.800} $\pm$ \textbf{0.018}		&3,328 & 72s\\
TT-LSTM			& 0.755 $\pm$ 0.033		& 3,392 & 81s\\ \hline
\end{tabular}
\end{small}
\end{center}
\end{table}

As expected, the baseline of TT-MLP model tends to perform well on the simpler video clips where the position of the face remains less changed over time, and can even outperform the plain GRU and LSTM. The TT-GRU and TT-LSTM, on the other hand, provide accuracy very close to the best state-of-the-art model (Tab. \ref{tab:ycf_other}) using Mean Sequence Sparse Representation-based Classification \cite{ortiz2013face} as feature extraction. 

\begin{table}[h]
\caption{State-of-the-art Results on Youtube Celebrities Face Dataset, in comparison with our best model. }
\label{tab:ycf_other}
\begin{center}
\begin{small}
\begin{tabular}{| l  | c | } \hline
Original: \cite{kim2008face} 				& 0.712 \\ \hline \hline
\cite{harandi2013dictionary} 	& 0.739 \\
\cite{ortiz2013face} 				& \textbf{0.808} \\
\cite{faraki2016image}			& 0.728 \\ \hline \hline 
Our best model (TT-GRU) 		& 0.800 \\ \hline
\end{tabular}
\end{small}
\end{center}
\end{table}

\paragraph{Experimental Settings} \mbox{} \\
We applied 0.25 Dropout \cite{srivastava2014dropout} for both input-to-hidden and hidden-to-hidden mappings in plain GRU and LSTM as well as their respective TT modifications; and 0.01 ridge regularization for the single-layered classifier. The models were implemented in Theano \cite{Bastien-Theano-2012} and deployed in Keras \cite{keras2015}. We used the Adam \cite{kingma2014adam} step rule for the updates with an initial learning rate 0.001.

\section{Conclusions and Future Work} \label{sec:conclusion}
We proposed to integrate Tensor-Train Layers into Recurrent Neural Network models including LSTM and GRU,  which enables them to be trained end-to-end on high-dimensional sequential data.
We tested such integration on three large-scale realistic video datasets. In comparison to the plain RNNs, which performed very poorly on these video datasets, we could empirically show that the integration of the Tensor-Train Layer alone significantly improves the modeling performances. In contrast to related works that heavily rely on deep and large CNNs, one advantage of our classification model is that it is simple and lightweight, reducing the number of free parameters from tens of millions to thousands. This would make it possible to train and deploy such models on commodity hardware and mobile devices. On the other hand, with significantly less free parameters, such tensorized models can be expected to be trained with much less labeled data, which are quite expensive in the video domain. 

More importantly, we believe that our approach opens up a large number of possibilities to model high-dimensional sequential data such as videos using RNNs directly. 
In spite of its success in modeling other sequential data such as natural language, music data etc., RNNs have not been applied to video data in a fully end-to-end fashion, presumably due to the large input-to-hidden weight mapping.
With TT-RNNs that can directly consume video clips on the pixel level, many RNN-based architectures that are successful in other applications, such as NLP, can be transferred to modeling video data: one could implement an RNN autoencoder that can learn video representations similar to \cite{sr2015uns}, an Encoder-Decoder Network \cite{cho2014properties} that can generate captions for videos similar to \cite{donahue2015long}, or an attention-based model that can learn on which frame to allocate the attention in order to improve the classification.

We believe that the TT-RNN provides a fundamental building block that would enable the transfer of techniques from fields, where RNNs have been very successful, to fields that deal with very high-dimensional sequence data --where RNNs have failed in the past.

The source codes of our TT-RNN implementations and all the experiments in Sec. \ref{sec:experiments} are publicly available at \url{https://github.com/Tuyki/TT_RNN}. In addition, we also provide codes of unit tests, simulation studies as well as experiments performed on the HMDB51 dataset \cite{Kuehne11}. 



\bibliography{TTRNN_ICML2017}
\bibliographystyle{icml2017}

\end{document}